\documentclass{article}

\usepackage{PRIMEarxiv}

\usepackage[utf8]{inputenc} 
\usepackage[T1]{fontenc}    
\usepackage{hyperref}       
\usepackage{url}            
\usepackage{booktabs}       
\usepackage{amsfonts}       
\usepackage{nicefrac}       
\usepackage{microtype}      
\usepackage{lipsum}
\usepackage{fancyhdr}       
\usepackage{graphicx}       
\graphicspath{{media/}}     

\usepackage{natbib}
\usepackage{adjustbox}
\usepackage{bbm}
\usepackage{bm}
\usepackage{amsmath}
\usepackage{booktabs}

\usepackage{makecell}
\usepackage{caption}
\usepackage{subcaption}
\usepackage{multirow}
\usepackage{xurl}
\usepackage[dvipsnames]{xcolor}

\usepackage{microtype}

\usepackage{array}
\usepackage{enumitem}
\usepackage{bm}
\usepackage{amsmath}
\usepackage{booktabs}
\usepackage{algorithm}
\usepackage{algpseudocode}
\usepackage{amssymb}
\usepackage{pifont}
\usepackage{tabularx}
\newcolumntype{Y}{>{\centering\arraybackslash}X}

\title{AsyncET: Asynchronous Learning for Knowledge Graph Entity Typing with Auxiliary Relations}

\author{
  Yun-Cheng Wang \\
  University of Southern California \\
  Los Angeles, USA\\
  \texttt{yunchenw@usc.edu} \\
   \And
  Xiou Ge \\
  University of Southern California \\
  Los Angeles, USA\\
  \texttt{xiouge@usc.edu} \\
   \And
  Bin Wang \\
  National University of Singapore \\
  Singapore\\
  \texttt{bwang28c@gmail.com} \\
   \And
  C.-C. Jay Kuo \\
  University of Southern California \\
  Los Angeles, USA\\
  \texttt{cckuo@sipi.usc.edu} \\
}

\begin{document}
\maketitle

\begin{abstract}
Knowledge graph entity typing (KGET) is a task to predict the missing
entity types in knowledge graphs (KG).  Previously, KG embedding (KGE)
methods tried to solve the KGET task by introducing an auxiliary
relation, `hasType', to model the relationship between entities and
their types. However, a single auxiliary relation has limited
expressiveness for diverse entity-type patterns. We improve the
expressiveness of KGE methods by introducing multiple auxiliary
relations in this work.  Similar entity types are grouped to reduce the
number of auxiliary relations and improve their capability to model
entity-type patterns with different granularities.  With the presence of
multiple auxiliary relations, we propose a method adopting an
\textbf{Async}hronous learning scheme for \textbf{E}ntity
\textbf{T}yping, named AsyncET, which updates the entity and type
embeddings alternatively to keep the learned
entity embedding up-to-date and informative for entity type prediction.
Experiments are conducted on two commonly used KGET datasets to show
that the performance of KGE methods on the KGET task can be
substantially improved by the proposed multiple auxiliary relations and
asynchronous embedding learning. Furthermore, our method has a
significant advantage over state-of-the-art methods in model sizes and
time complexity. 
\end{abstract}


\section{Introduction}\label{sec:intro}

\begin{figure}[t]
\centering
\includegraphics[width=0.75\textwidth]{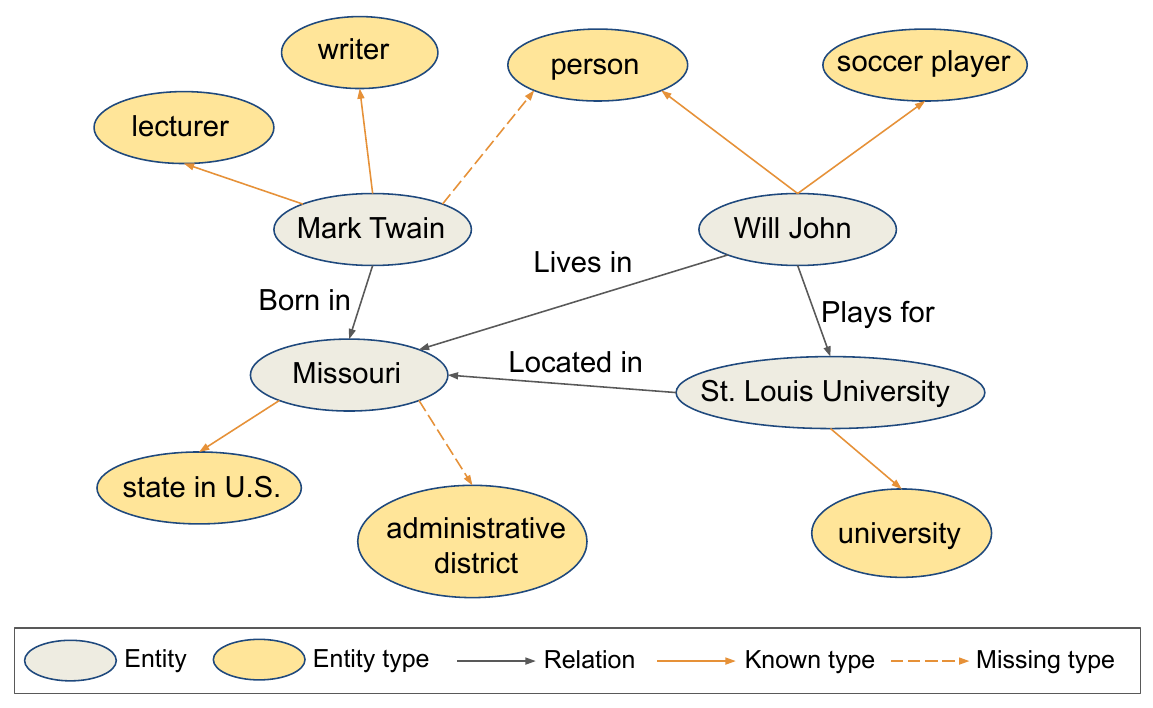}
\caption{An example KG with missing entity types.}\label{fig:example}
\end{figure}

Knowledge graph (KG) stores human-readable knowledge in a
graph-structured format, where nodes and edges denote entities and
relations, respectively. There are multiple relation types in KGs to
describe the relationship between two entities.  A (head entity,
relation, tail entity) factual triple is a basic component in KGs. In
addition to different relation types, each entity also comes with
multiple types to describe the high-level abstractions of an
entity\footnote{We refer to ``entity type" when using the term ``type"
in the remainder of this paper.}.  Fig.~\ref{fig:example} shows an
example KG containing entity type information.  As shown in the figure,
each entity can be labeled with multiple types.  For example, the entity
``\emph{Mark Twain}'' has types ``\emph{writer}'' and
``\emph{lecturer}'' at the same time. Entity types are crucial in
several artificial intelligence (AI) and natural language processing
(NLP) applications, such as drug discovery \citep{lin2017drug}, entity
alignment, \citep{xiang2021ontoea, huang2022cross}, and entity linking
\citep{gupta2017entity}.  In real-world applications, entity types could
be missing, e.g., having type ``\emph{writer}'' without type
``\emph{person}'', due to prediction errors from information extraction
models \citep{yaghoobzadeh2016noise, yaghoobzadeh2016corpus}.  Such
missing types can be inferred from the existing information in KG. For
example, in Fig. \ref{fig:example}, we can infer that ``\emph{Mark
Twain}'' has a missing type ``\emph{person}'' given that there is a known
type ``\emph{writer}'' and the relation ``\emph{born in}''.  Thus,
knowledge graph entity typing (KGET) is a task to predict the missing
types based on the observed types and triples in the KGs.  KGET
methods can serve as refinement mechanisms for real-world knowledge
bases.

Knowledge graph embedding (KGE) methods achieve great success in
predicting missing triples in KGs \citep{ji2021survey}, and they are
extended to solve the KGET task in \citep{moon2017learning}.  Since the
type labels are stored in the format of tuples (entity, type), an
auxiliary relation, \emph{hasType}, is first introduced to convert the
typing tuples into triples (entity, \emph{hasType}, type), and, then, a
KGE method \citep{moon2017context} is adopted to predict missing types.
Although such a method is time- and parameter-efficient, it does not
perform well since the relationship between entities and types
is too diverse to be modeled by a single relation.  In addition, such a
method does not consider the neighborhood information. This affects the
performance of entity type prediction as well.  Other methods are
proposed to improve the model's expressiveness.  Embedding-based methods,
such as ConnectE~\citep{zhao2020connecting}, first learn embeddings 
for entities and types separately using KGE methods. 
Then, a linear projection matrix is learned to minimize the
distance between the entity and the type spaces.  Another work leverages
multi-relational graph convolution networks (R-GCN)
\citep{schlichtkrull2018modeling} to encode the neighborhood information
into entity embeddings.  The attention mechanism is also explored in
\citep{zhao2022connecting} to control the contributions of neighbors when
predicting an entity type.  Afterward, multi-layer perceptrons (MLPs)
are cascaded to predict the entity types based on the learned entity
embeddings. The KGET task is, therefore formulated as a multi-label
classification problem. Although GCN-based methods offer superior
performance, they are not applicable in a resource-constrained environment, 
such as mobile/edge devices and real-time prediction \citep{kuo2022green}, 
due to their high inference time complexity and large model sizes. 
In addition, training GCNs could be time- and
memory-consuming. Their applicability to large-scale KGs is challenging.
It is desired to develop a KGET method of low inference time complexity,
small model sizes, and good performance. This is the objective of this
work. 

\begin{figure}[t]
\centering
\includegraphics[width=0.75\textwidth]{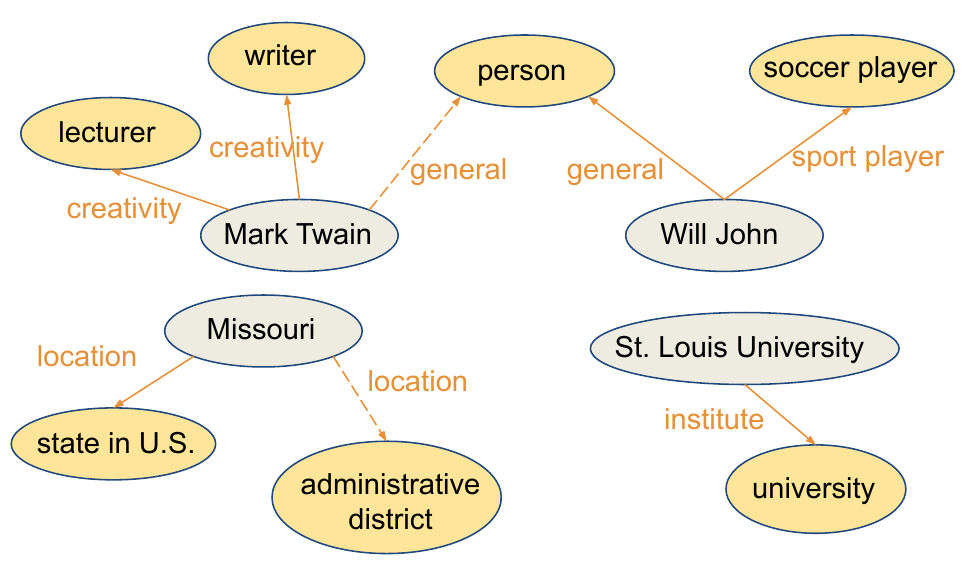}
\caption{Illustration of using multiple auxiliary relations to 
model the relationship between entities and entity types.}\label{fig:aux}
\end{figure}

As argued above, a single auxiliary relation is not sufficient to model
the relationship between entities and types.  Here, we introduce more
auxiliary relations to improve the expressiveness of KGE models on the
entity-type prediction task.  This idea is illustrated in Fig.
\ref{fig:aux}, where we show an example of using multiple auxiliary
relations to model the entity-type relationship.  It is intuitive that
we should use different auxiliary relations to model typing relationship
for type ``\emph{administrative district}'' and type ``\emph{person}''
since they describe two different concepts of entities.  Similarly, type
``\emph{writer}'' and type ``\emph{soccer player}'' should adopt
different auxiliary relations since they are semantically different.
They should not be close to each other in the embedding space.  On the
other hand, for other types, such as ``\emph{writer}'' and
``\emph{lecturer}'', they co-occur with each other more frequently.
Thus, they can adopt the same auxiliary relation for model simplicity.

Along this direction, we introduce multiple auxiliary relations based on
the ``context'' of types. The context of a type is defined as a
collection of attributes of its entities. It can be viewed as a discrete
representation of the type.  As such, the local KG structure is
implicitly encoded when auxiliary relations are used. Next, we propose a
method adopting an \textbf{Async}hronous learning scheme for
\textbf{E}ntity \textbf{T}yping, named AsyncET, to obtain better
embeddings for entities and types for the entity type prediction task.
The training process consists of two stages: 1) link prediction, and 2)
type prediction.  Entity embeddings are first optimized by training with
only factual triples to predict missing links.  Then, the typing
information is used to train type embeddings and fine-tune entity
embeddings by predicting missing types.  Two training stages alternate
as the training progresses. The asynchronous training schema keeps the
learned entity embedding up-to-date and informative for entity type
prediction. Experiments conducted on two KGET datasets demonstrate that
the
proposed multiple auxiliary relations and asynchronous training framework can substantially improve the performance of the KGET task.  Furthermore, AsyncET has a significant advantage over
existing KGET methods in model sizes and time complexity. 

The main contributions of this paper are summarized below. 
\begin{itemize}
\item We introduce a novel strategy, called multiple auxiliary
relations, to model relationships between entities and types. 
\item We propose a new asynchronous embedding learning framework, named
AsyncET, to obtain better entity and type embeddings for the KGET task. 
\item AsyncET can improve the performance of KGE models on the KGET
task substantially while being more efficient in inference time and
model size. 
\end{itemize}

\section{Related Work}\label{sec:review}

\subsection{Embedding-based Methods} 

KGE methods rely on an auxiliary relation, \emph{hasType}, to form typing
triples (entity, \emph{hasType}, type) so as to solve the KGET task.
Synchronous training is often adopted to mix the factual triples
and typing triples when training embeddings.  Such methods have 
advantages in inference and model parameter efficiency.  However, the
performance is difficult to improve due to overly simplifying the
relationship between entities and types.  Other embedding-based methods
tend to learn an entity space and a type space separately. Then, a
mapping between two vector spaces is learned to predict the missing
connections between entities and types.  ETE \citep{moon2017learning}
tried to minimize the distance between the learned entity and type space
through the L1-norm.  ConnectE \citep{zhao2020connecting} adopts a linear
projection matrix to connect entity and type embedding spaces. JOIE
\citep{hao2019universal} proposes two training objectives, cross-view
grouping and cross-view transformation, to ensure entities with
similar types are closely embedded. TransC \citep{lv2018differentiating}
encodes entities and types in the same embedding space as
high-dimensional balls. Several constraints are imposed to maintain the
hierarchy among entity types.  CORE \citep{ge2021core} learns KGE in a
complex subspace \citep{sun2018rotate, trouillon2016complex} for entities
and types individually. Then, a linear regression problem is solved to
link entities with their corresponding types.  Although the
embedding-based methods generally contain fewer model parameters and
have lower inference time, their performance highly depends on the
expressiveness of the model used to describe the entity-type relations.

\subsection{Deep Neural Network Methods} 

The neighborhood information is important in the KGET task since the
types of an entity can often be determined by the neighboring entities
and types.  Following this line of thought, multi-relational GCNs
\citep{schlichtkrull2018modeling, zhao2021wgcn, vashishth2019composition}
are proposed for the KGET task.  First, entity embeddings are aggregated
from the neighboring entities and types in GCNs.  Then, multi-layer
perceptrons (MLPs) are used to predict missing entity types by solving a
multi-label classification problem.  Since not all neighbors contribute
to entity type prediction equally, an attention mechanism has been used to
achieve better performance in recent work.  For example, ConnectE-MRGAT
\citep{zhao2022connecting} adopts graph attention networks (GATs)
\citep{velivckovic2017graph, nathani2019learning} to solve the KGET task.
CET \citep{pan2021context} uses two attention mechanisms (i.e., N2T and
Agg2T) to aggregate the neighborhood information.  AttET
\citep{zhuo2022neighborhood} adopt a type-specific attention mechanism to
improve the quality of entity embeddings.  TET \citep{hu2022transformer}
uses a transformer as the entity encoder to aggregate the neighboring
information.  Although deep neural network methods can achieve superior
performance, their inference complexity and model sizes are much larger
than those of embedding-based methods. 

\section{Methodology}\label{sec:method}

\subsection{Notations}

We use $\mathcal{G}$ to denote a KG containing a collection of factual
triples; namely,
\begin{equation}
\mathcal{G} = \{(e^{head}, r, e^{tail}) \mid e^{head}, e^{tail} \in \mathcal{E}, 
r \in \mathcal{R}\},
\end{equation}
where $\mathcal{E}$ and $\mathcal{R}$ represent sets of entities and relations in
the KG, respectively. The type information is denoted as
\begin{equation}
\mathcal{I} = \{(e, t) \mid e \in \mathcal{E}, t \in \mathcal{T}\},
\end{equation}
where $\mathcal{T}$ is a set of entity types in the KG.  In order to
group similar types based on the attributes of their associated
entities, we define the context of type $t$ as a collection of
relations that co-occur with entities of type $t$
\begin{equation}
\mathcal{C}_t = \{r \mid (e^{head}, r, e^{tail}) \in \mathcal{G}, (e^{head}, t) 
\in \mathcal{I}\}.
\end{equation}
For example, the context of type \emph{person} is \{\emph{Born in},
\emph{Lives in}, \emph{Plays for}, ... \}. It contains all attributes
an entity of type \emph{person} can have.  The context of type $t$ can
be seen as a discrete representation for $t$ that encodes the local
structure of the KG. 

\subsection{Auxiliary Relations} \label{sec:aux}

Previous KGE models have only one auxiliary relation - \emph{hasType}.
They convert a typing tuple, $(e, t)$, into a typing triple, $(e,
\textit{hasType}, t)$.  However, a single relation is not sufficient to
model diverse entity-type patterns. Here, we aim to find a mapping such
that, given entity type $t$, an auxiliary relation $p = Aux(t)$ is
assigned to form new typing triples $(e, p, t)$, where $t \in
\mathcal{T}$, $p \in \mathcal{P}$, and $\mathcal{P}$ denotes a set of
auxiliary relations in the KG.  The objective is to maximize the
capabilities of auxiliary relations to model every entity-type
relationship. We compare three methods for the design of auxiliary
relations below. 

\textbf{Bijective assignment.} A straightforward solution to enhance
the expressiveness of auxiliary relations is to assign a unique auxiliary
relation to each type, called the bijective assignment.  It can model
diverse typing patterns well by exhaustively modeling every possible
typing pattern in the KG.  However, when the KG contains a large number
of types, this assignment has several shortcomings. First, the model
optimization is less stable since the number of model parameters
increases significantly. Second, it is too fine-grained to perform well
on the test dataset.  Third, its inference time is much longer.
Therefore, it is essential to group similar types and assigns auxiliary
relations to each group of types. 

\textbf{Taxonomy-based assignment.} Taxonomy is a hierarchical
organization of concepts in KGs.  For example, type ``/film/producer" in
Freebase \citep{bollacker2008freebase} is a type under category ``film"
with the subcategory ``producer". To group types based on the taxonomy, we
can group them based on the first layer of the taxonomy. For instance,
``/film/producer" will belong to the ``film" group, and
``/sports/sports\_team" will belong to the ``sports" group.  However,
such a taxonomy might not be available for some KGs, say, YAGO subsets
\citep{mahdisoltani2014yago3}. Furthermore, the
first-layer-taxonomy-based assignment may not have enough granularity
for some types.  To address these issues, we propose a new assignment
method below. 

\textbf{Efficient assignment.} To strike a balance between a small
number of auxiliary relations, $|\mathcal{P}|$, and high expressiveness
of entity-type modeling, we maximize similarities among the types in the
same group and minimize similarities among different groups.
Mathematically, we adopt the Jaccard similarity between the type contexts
to define similarities between types. It can be written in the form of
\begin{equation}\label{equ:sim}
Sim(t, t') = \frac{|\mathcal{C}_{t} \cap \mathcal{C}_{t'}|}{|\mathcal{C}_{t} 
\cup \mathcal{C}_{t'}|}.
\end{equation}

\renewcommand{\algorithmicrequire}{\textbf{Initialization:}}
\renewcommand{\algorithmicensure}{\textbf{Iteration:}}

\begin{figure}[t]
  \centering
  \begin{minipage}{0.75\linewidth}
    \begin{algorithm}[H]
    \caption{Find anchors for efficient auxiliary relation assignment} \label{alg:aux}
        \begin{algorithmic}
        \Require \\\noindent The uncovered relations in the KG: $\mathcal{U} = \mathcal{R}$ 
        \\\noindent The set of existing anchor types: $\mathcal{A} = \varnothing$
        \Ensure 
        \While{$\mathcal{U} \neq \varnothing$}
            \State $t = \operatorname*{argmax}_{t \in \mathcal{T}} |\mathcal{C}_t \cap \mathcal{U}|$
            \State $\mathcal{U} \gets \mathcal{U} \setminus \mathcal{C}_t$
            \State $\mathcal{A} \gets \mathcal{A} \cup \{t\}$ 
        \EndWhile
        \end{algorithmic}
    \end{algorithm}
  \end{minipage}
\end{figure}

Then, based on the well-defined similarity function between types, 
grouping similar types with the minimum number of groups 
can be formulated as a min-max optimization problem.  
Such an optimization problem is NP-Hard. Here, we develop a
greedy algorithm to find an approximate optimal assignment as elaborated
in the following. First, we identify several anchor types
to be the centroid of each group. Initially, the anchor type set is empty,
and all types are not covered.  The method iteratively selects the type,
which has not yet been selected, with the largest intersection of
context and the uncovered relations $|\mathcal{C}_t \cap \mathcal{U}|$
as a new anchor. The iteration ends when the union of all anchors'
contexts is equal to $\mathcal{R}$. The process of finding anchor types
is depicted in Algorithm \ref{alg:aux}. Then, we assign a unique auxiliary
relation to each anchor type.  The non-anchor types will find their most
similar anchor types and share the same auxiliary relations with the
anchor type. 

To verify whether the proposed algorithm can generate reasonable
results, we show examples of the grouped entity types in
Table~\ref{tab:aux_example}.  We see from the table that the auxiliary
relation \# 2 is assigned to types of persons who work in the
entertainment industry, auxiliary relation \# 20 is assigned to types
that are mostly sports teams, and auxiliary relation \# 27 is assigned
to geographical locations.  Similar types are successfully grouped
together while types of distant semantic meanings are separated. 

\begin{table}
\centering
\caption{Examples of auxiliary relations and the corresponding 
entity types using the proposed efficient assignment,
where anchor types are marked in boldface.}\label{tab:aux_example}
\begin{tabular}{ c l }
\toprule
Auxiliary Relation & Entity Types \\
\midrule
\multirow{4}{*}{\# 2} & \textbf{/film/producer} \\
  & /TV/tv\_director \\
  & /film/writer \\
  & /film/film\_story\_contributor\\
\midrule
\multirow{4}{*}{\# 20} & \textbf{/sports/sports\_team} \\
   & /soccer/football\_team \\
   & /baseball/baseball\_team \\
   & /sports/school\_sports\_team \\
\midrule
\multirow{4}{*}{\# 27} & \textbf{/location/administrative\_division} \\
   & /film/film\_location \\
   & /fictional\_universe/fictional\_setting \\
   & /location/citytown \\
\bottomrule
\end{tabular}
\end{table}

\begin{figure*}[t]
\centering
\includegraphics[width=\textwidth]{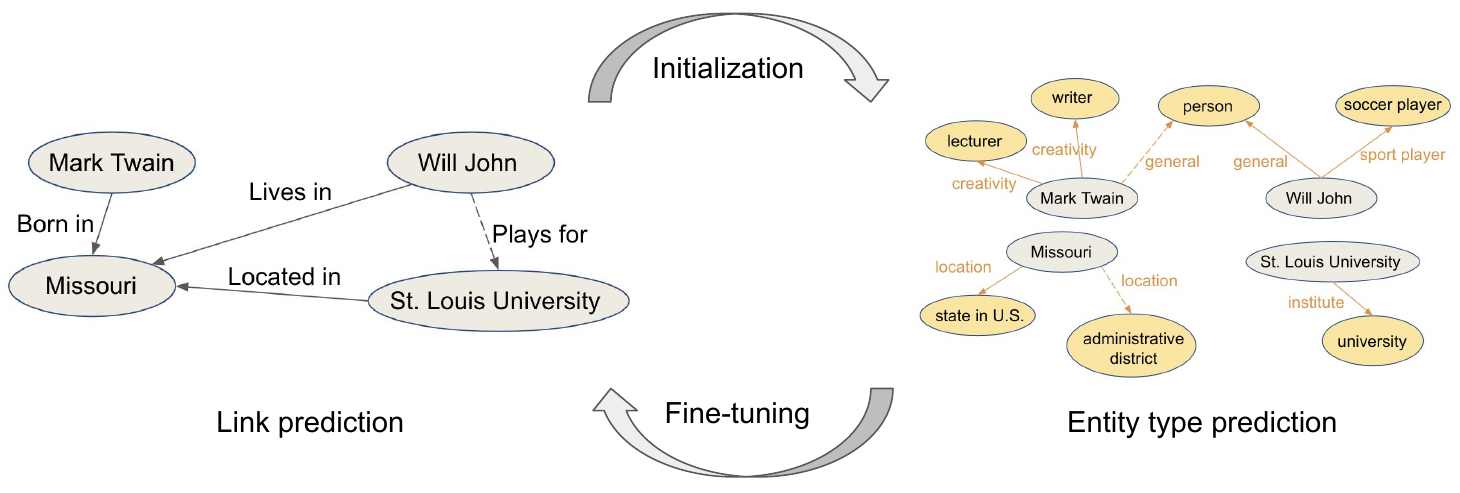}
\caption{A diagram of the training process in AsyncET.}\label{fig:asyncet}
\end{figure*}

\subsection{\textbf{AsyncET}: \textbf{Async}hronous Embedding Learning for 
Knowledge Graph \textbf{E}ntity \textbf{T}yping}

After auxiliary relations are defined, typing tuples $(e, t)$ can 
be converted into typing triples $(e, p, t)$. Such typing triples form a typing graph
\begin{equation}
    \mathcal{TG} = \{(e, p, t) \mid (e, t) \in \mathcal{I}, p = Aux(t)\}.
\end{equation}
Instead of mixing the original triples and the newly constructed typing
triples together in embedding learning, we optimize the entity and type
embeddings on the original KG $\mathcal{G}$ and the typing graph
$\mathcal{TG}$ alternatively.  That is, the embedding learning process
is divided into two stages.  In stage 1, the entity embeddings are
trained on $\mathcal{G}$ using a link prediction task.  The learned
entity embeddings serve as an initialization for embedding learning in
stage 2, where we use the typing graph $\mathcal{TG}$ to learn type 
embeddings and fine-tune entity embeddings with typing triples. The two training
stages are optimized alternatively. Fig.  \ref{fig:asyncet} illustrates the
training process of asynchronous embedding learning.  Details of
each training stage are elaborated below. 

\textbf{Stage 1: Link prediction.} The goal of this stage is to obtain a
good initialization of entity embeddings that can be used to predict the
missing types. We follow the training loss in \citep{sun2018rotate} with
the self-adversarial negative sampling.  The link prediction loss is
defined as
\begin{equation}\label{equ:lp_loss}
\begin{split}
        \mathcal{L}_{lp} = {} & - \log(\sigma(f(\bm{e^{head}}, \bm{r}, \bm{e^{tail}}))) \\
                         & - \sum_{i=1}^n p(e'_i, r, e''_i) 
                    \log(\sigma( - f(\bm{e'}_i, \bm{r}, \bm{e''}_i))), 
\end{split}
\end{equation}
where $(e'_i, r, e''_i)$ is the negative samples generated by corrupting the head
and tail entities, $f(\bm{e^{head}}, \bm{r}, \bm{e^{tail}})$ is the
scoring function in the KGE model, and $\bm{e^{head}}, \bm{r}, \bm{e^{tail}}$
are embeddings for the head entity, relation, and tail entity, respectively.
The self-adversarial negative sampling distribution is defined as
\begin{equation} \label{equ:self-adv}
        p(e'_j, r, e''_j) = \frac{\exp ( \alpha f(\bm{e}'_j, \bm{r}, \bm{e}''_j) )}
{\sum_{i=1}^n \exp ( \alpha f(\bm{e}'_i, \bm{r}, \bm{e}''_i) )},
\end{equation}
where $\alpha$ is the temperature to control the smoothness of the softmax function.
As a result, negative samples with lower scores are assigned smaller
weights for optimization as they are well-trained already, and the model
can focus on optimizing the hard cases. 

\textbf{Stage 2: Entity type prediction.} In this stage, we fine-tune the 
entity embeddings and train the type embeddings using only typing 
triples. We adopt a loss similar to (\ref{equ:lp_loss}) to predict the 
missing entity types.
\begin{equation}\label{equ:tp_loss}
\begin{split}
        \mathcal{L}_{tp} = {} & - \log(\sigma(f(\bm{e}, \bm{p}, \bm{t}))) \\
                         & - \sum_{i=1}^n p(e, p'_i, t'_i) 
                    \log(\sigma( - f(\bm{e}, \bm{p}'_i, \bm{t}'_i))), 
\end{split}
\end{equation}
where $(e, p'_i, t'_i)$ is a negative sample for entity type prediction
generated by replacing the valid types with a random type $t'_i$ and 
the corresponding auxiliary relation $p'_i$.
The auxiliary relations are assigned based on mappings, $p = Aux(t)$ 
and $p'_i = Aux(t'_i)$. Since the number of entity types is much fewer
than the number of entities (i.e. $|\mathcal{T}| << |\mathcal{E}|$), 
false negatives, (i.e. $(e, p'_i, t'_i) \in \mathcal{TG}$) 
are more prevalent for entity-type prediction. To address this issue,
we adopt false-negative-aware negative sampling distribution 
introduced in \citep{pan2021context}. It can be written as
\begin{equation} \label{equ:fna}
p(e, p'_j, t'_j) = x - x^2,
\end{equation}
where
\begin{equation}
x = \sigma( - f(\bm{e}, \bm{p}'_i, \bm{t}'_i)).
\end{equation}
Then, negative samples with the highest scores are assigned lower
weights as they are possibly false negatives. Similar to the
self-adversarial loss, negative samples with the lowest scores are
already well-trained, so they are assigned smaller negative sampling
weights. 

\begin{table}
\centering
\caption{Dataset statistics.}
\begin{tabular}{ c l cc }
\toprule
&& FB15k-ET & YAGO43k-ET \\
\midrule
\multirow{3}{*}{$\mathcal{G}$} & \# entities &14,951 &42,334 \\
&\# relations &1,345 &37 \\
&\# triples &483,142 &331,686 \\
\midrule
\multirow{7}{*}{$\mathcal{TG}$} &\# types &3,584 &45,182 \\
&\# train &136,618 &375,853 \\
&\# valid &15,848 &43,111 \\
&\# test  &15,847 &43,119 \\
\cmidrule{2-4}
&\# $p$-bijective  &3,584 &45,182 \\
&\# $p$-taxonomy   &89 &-  \\
&\# $p$-efficient  &54 &10 \\
\bottomrule
\end{tabular}
\label{tab:dataset}
\end{table}

\begin{table*}[t]
\centering
\caption{Results on KGET datasets, where the best performance in each column is shown
in boldface, and the second-best performance is underlined.}
\begin{tabular}{ l cccc cccc }
\toprule
\multirow{2}{*}{Models} & \multicolumn{4}{c}{FB15kET} & \multicolumn{4}{c}{YAGO43kET} \\
\cmidrule(l){2-5} \cmidrule(l){6-9}
& MRR & H@1 & H@3 & H@10 & MRR & H@1 & H@3 & H@10 \\
\midrule
TransE \citep{bordes2013translating}& 0.618 & 0.504 & 0.686 & 0.835 & 0.427 & 0.304 & 0.497 & 0.663 \\
RotatE \citep{sun2018rotate}        & 0.632 & 0.523 & 0.699 & 0.840 & 0.462 & 0.339 & 0.537 & 0.695 \\
CompoundE \citep{ge2022compounde}   & 0.640 & 0.525 & 0.719 & 0.859 & 0.480 & 0.364 & 0.558 & 0.703 \\
\midrule
ETE \citep{moon2017learning}        & 0.500 & 0.385 & 0.553 & 0.719 & 0.230 & 0.137 & 0.263 & 0.422 \\
ConnectE \citep{zhao2020connecting} & 0.590 & 0.496 & 0.643 & 0.799 & 0.280 & 0.160 & 0.309 & 0.479 \\
CET \citep{pan2021context}          & \textbf{0.697} & \textbf{0.613} & \underline{0.745} & 0.856 & \textbf{0.503} & \textbf{0.398} & \underline{0.567} & 0.696 \\
ConnectE-MRGAT \citep{zhao2022connecting}   & 0.630 & 0.562 & 0.662 & 0.804 & 0.320 & 0.243 & 0.343 & 0.482 \\
AttET \citep{zhuo2022neighborhood}  & 0.620 & 0.517 & 0.677 & 0.821 & 0.350 & 0.244 & 0.413 & 0.565 \\
\midrule 
AsyncET-TransE (Ours)                 & 0.659 & 0.552 & 0.729 & 0.859 & 0.452 & 0.341 & 0.518 & 0.684 \\
AsyncET-RotatE (Ours)                 & 0.668 & 0.564 & 0.735 & \underline{0.864} & 0.471 & 0.359 & 0.556 & \underline{0.717} \\
AsnycET-CompoundE (Ours)              & \underline{0.688} & \underline{0.581} & \textbf{0.755} & \textbf{0.885} & \underline{0.492} & \underline{0.380} & \textbf{0.574} & \textbf{0.721} \\
\bottomrule
\end{tabular}
\label{tab:main}
\end{table*}

\section{Experiments}\label{sec:experiment}

\subsection{Experimental Setup}

{\bf Datasets.} We adopt two KGET datasets, FB15k-ET and YAGO43k-ET
\citep{moon2017learning}, for evaluation.  FB15k-ET is derived from a
link prediction dataset, FB15K \citep{bordes2013translating}, extracted
from Freebase \citep{bollacker2008freebase} by adding typing tuples.
Freebase contains general relations between real-world entities.
YAGO43k-ET is derived from another link prediction dataset, YAGO43k
\citep{moon2017context}, extracted from YAGO \citep{mahdisoltani2014yago3}
by adding typing tuples. There are mostly attributes and relations for
persons in YAGO. The dataset statistics are summarized in Table
\ref{tab:dataset}, where $p$-bijective, $p$-taxonomy, and $p$-efficient
denote the auxiliary relations obtained from the bijective assignment,
the taxonomy-based assignment, and the efficient assignment described in
Sec. \ref{sec:aux}, respectively. Only FB15k-ET contains taxonomy labels
for the types. 

{\bf Implementation details.} We select three representative KGE methods
as the scoring functions to evaluate the effectiveness of AsyncET.
Specifically, we select TransE \citep{bordes2013translating}, RotatE
\citep{sun2018rotate}, and CompoundE \citep{ge2022compounde}.  TransE
models relations as translation in the vector space. It has several
limitations in expressiveness since it does not model symmetric relations
well. RotatE models relations as rotation in the complex embedding
space. CompoundE is a recently proposed KGE method that generalizes the
majority of distance-based methods by including compounding geometric
transformations such as translation, rotation, and scaling.  It also
operates in the complex embedding space. The scoring functions of three
KGE methods for AsyncE are written below.
\begin{itemize}
\item TransE~\citep{bordes2013translating}:
$$f
(\bm{e^{head}}, \bm{r}, \bm{e^{tail}}) = \gamma - \|\bm{e^{head}}+\bm{r}-\bm{e^{tail}}\|,
$$
where $\gamma$ is the margin. It's a hyperparameter that can be tuned.
\item RotatE~\citep{sun2018rotate}:
$$
f(\bm{e^{head}}, \bm{r}, \bm{e^{tail}}) = \gamma - \|\bm{e^{head}} \circ \bm{r} - \bm{e^{tail}} \|,
$$
where $\circ$ denotes the rotation operation in the complex embedding space.
\item CompoundE~\citep{ge2022compounde}:
$$
f(\bm{e^{head}}, \bm{r}, \bm{e^{tail}}) = \gamma - \|\mathbf{T_r\cdot R(\theta_r)\cdot 
S_r\cdot \bm{e^{head}} - \bm{e^{tail}} }\|,
$$
where $\mathbf{T_r}$, $\mathbf{R(\theta_r)}$, $\mathbf{S_r}$ denote the
translation, rotation, and scaling operations in CompoundE, respectively. 
\end{itemize}

For both datasets, we select the best hyper-parameters from a certain
search space under the embedding dimension $d = 500$ based on the
performance of the validation set for entity type prediction. The
search space is given below: 
\begin{itemize}
\item Number of negative samples $n_{neg} \in \{128, 256_{\star\diamond}, 512\}$; 
\item Learning rate $lr \in \{0.01_{\diamond}, 0.001_{\star}, 0.0001\}$; 
\item Softmax temperature $\alpha \in \{0.5, 1.0_{\star\diamond}\}$; 
\item Margin $\gamma \in \{8.0_{\star}, 12.0, 16.0, 20.0_{\diamond}\}$.  
\end{itemize}
The hyper-parameter settings adopted for FB15k-ET and YAGO43k-ET are
marked with $\star$ and $\diamond$, respectively. We also conduct
an ablation study on the performance against the number of alternate steps
between two training stages in asynchronous embedding learning in Sec.
\ref{sec:alter_round}.  Based on the study, we alternate two training
stages every 16 steps for both datasets. All experiments are conducted
using one NVIDIA Tesla P100 GPU. 

{\bf Evaluation metrics.} For the KGET task, the goal is to predict the
missing types given an entity, i.e., $(e, ?)$.  However, in AsyncET, we
convert the tuples into triples so the test queries become $(e, ?, ?)$.
Since each entity type is only modeled by one auxiliary relation, we
evaluate the joint plausibility $f(e, p', t')$, $\forall t' \in
\mathcal{T}$, where $p' = Aux(t')$ for a given query.  The valid entity
types should be ranked as high as possible compared to all other
candidates. Following the convention in \citep{bordes2013translating}, we
adopt the filtered setting, where all entity types in the KG serve as
candidates except for those observed ones.  Several commonly used
ranking metrics are adopted, including the Mean Reciprocal Rank (MRR) and
Hits@k (k=1, 3, and 10). 

\begin{table*}[t]
\setlength\tabcolsep{8pt}
\centering
\caption{Ablation study on asynchronous representation learning and
different auxiliary relations. The MRR performance is reported. The best performance
in each column is shown in boldface.}\label{tab:ablation}
\begin{tabularx}{\textwidth}{ cc *{6}{Y} }
\toprule
&& \multicolumn{3}{c}{FB15kET} & \multicolumn{3}{c}{YAGO43kET} \\
\cmidrule(l){3-5} \cmidrule(l){6-8}
\emph{Training} & \emph{Aux. Rel.} & TransE & RotatE & CompoundE & TransE & RotatE & CompoundE \\
\midrule
Syn. & \emph{hasType} &0.618 &0.632 &0.640 &0.427 &0.462 &0.480 \\
Syn. & $p$-bijective  &0.532 &0.534 &0.581 &0.362 &0.388 &0.407 \\
Syn. & $p$-taxonomy   &0.545 &0.550 &0.603 &-     &-     &-     \\
Syn. & $p$-efficient  &0.565 &0.564 &0.625 &0.418 &0.438 &0.455 \\
\midrule
Asyn. & \emph{hasType} &0.621 &0.624 &0.638 &0.443 &0.458 &0.474 \\
Asyn. & $p$-bijective  &\textbf{0.659} &\textbf{0.668} &\textbf{0.688} &0.391 &0.418 &0.442 \\
Asyn. & $p$-taxonomy   &0.633 &0.641 &0.664 &-     &-     &-     \\
Asyn. & $p$-efficient  &0.654 &0.661 &0.682 &\textbf{0.452} &\textbf{0.471} &\textbf{0.492} \\
\bottomrule
\end{tabularx}
\end{table*}

\subsection{Main Results}

The experimental results on two KGET datasets are given in Table
\ref{tab:main}. Models are clustered into three groups: 1) KGE methods
trained using a single auxiliary relation (i.e., \emph{hasType}), 2)
other type-embedding methods and models using graph neural networks, 3)
proposed AsyncET with TransE, RotatE, and CompoundE scoring functions.
For group 3, we report the best performance using $p$-bijective,
$p$-taxonomy, or $p$-efficient.  Detailed comparison of the effects of
different auxiliary relations will be discussed in Sec. \ref{sec:ablation}. 
We have the following observations from the table.  AsyncET is
significantly better than KGE methods trained with only one auxiliary
relation.  CET performs better in exact matches (i.e. H@1) than AsyncET
since they adopt a graph neural network to learn entity embeddings from
neighbors. However, the performance of some type-embedding methods, such
as ETE and ConnectE, is much worse than that of AsyncET for both
datasets since they do not encode the neighboring information
effectively.  Recent methods, such as CET, ConnectE-MRGAT, and AttET,
adopt attention mechanisms to obtain context-aware entity embeddings to
predict the missing types.  AsyncET encodes entities' attributes
through auxiliary relations and asynchronous training.
AsyncET using scoring functions of TransE, RotatE, and CompoundE
outperforms attention-based methods in all metrics except for CET.
Furthermore, AsyncET-CompoundE can even outperform CET in H@3 and H@10. 

\begin{figure}[t]
\centering
\includegraphics[width=0.6\textwidth]{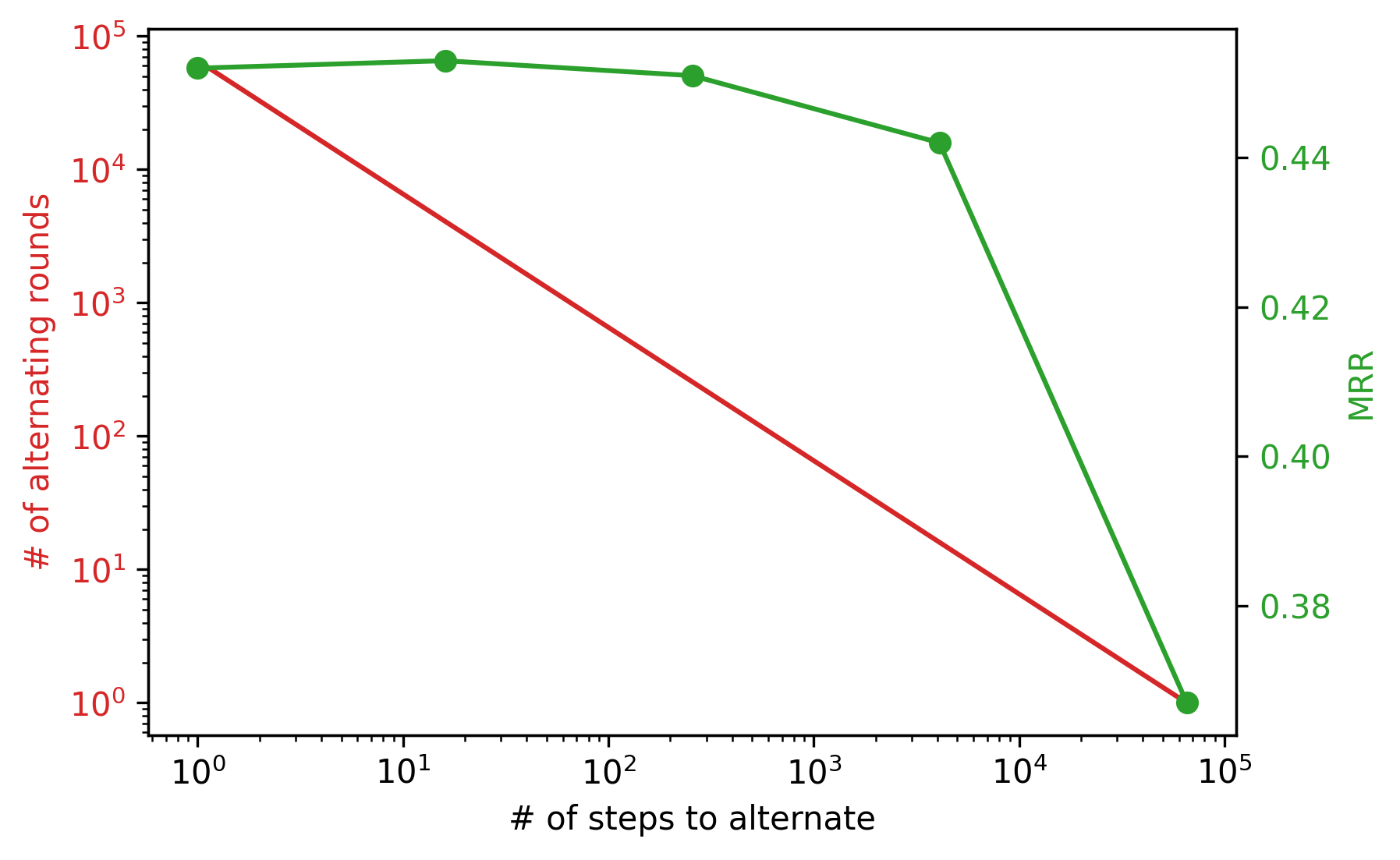}
\caption{The MRR performance as a function of the number of alternating 
rounds between two stages in asynchronous representation learning.}
\label{fig:alternate}
\end{figure}

\subsection{Ablation Study} \label{sec:ablation}

To analyze the effectiveness of asynchronous training and
different auxiliary relation assignments, we conduct an ablation study in
Table~\ref{tab:ablation}, where the MRR performance of several methods
for two datasets is reported. We compare the following:
\begin{itemize}
\item Synchronous vs. asynchronous training;
\item Single auxiliary relation \emph{hasType} vs. multiple auxiliary relations 
with $p$-bijective, $p$-taxonomy, and $p$-efficient designs;
\item TransE, RotatE, or CompoundE scoring functions.
\end{itemize}

As shown in the table, asynchronous training consistently outperforms
synchronous training.  The performance improvement of asynchronous
training when using only one auxiliary relation, \emph{hasType}, is not
significant since the second stage of the embedding learning is trained
on a single-relational graph. When there are multiple auxiliary
relations, mixing typing triples with original factual triples in
embedding training using the synchronous framework still yields poor
performance. This could be attributed to the fact that KGE methods 
are difficult to train when there are too many relations. 
The performance improves significantly when we decompose 
the training process into two stages under the asynchronous framework. 

When using multiple auxiliary relations to model the typing
relationship, the bijective assignment works well in the dataset with
fewer entity types, e.g.  FB15k-ET.  However, it performs worse than the
efficient assignment on the dataset with more entity types, e.g.
YAGO43kET.  In datasets with many entity types, the bijective assignment
introduces larger amounts of new parameters, making the embedding
training more difficult to converge.  The efficient assignment can
achieve the best performance on YAGO43kET, showing that such an
assignment can capture the context in the KG and similarities among
entity types well.  Surprisingly, the taxonomy-based assignment does not
perform well on both datasets. One possible reason is that grouping
entity types based on only the first layer of the taxonomy ignores the
granularities of different entity-type patterns.  Such a problem might
be solved by considering more layers in the taxonomy.

\begin{figure*}[t]
     \centering
  \begin{tabular}{c@{}c@{}c}
    \includegraphics[width=.32\linewidth]{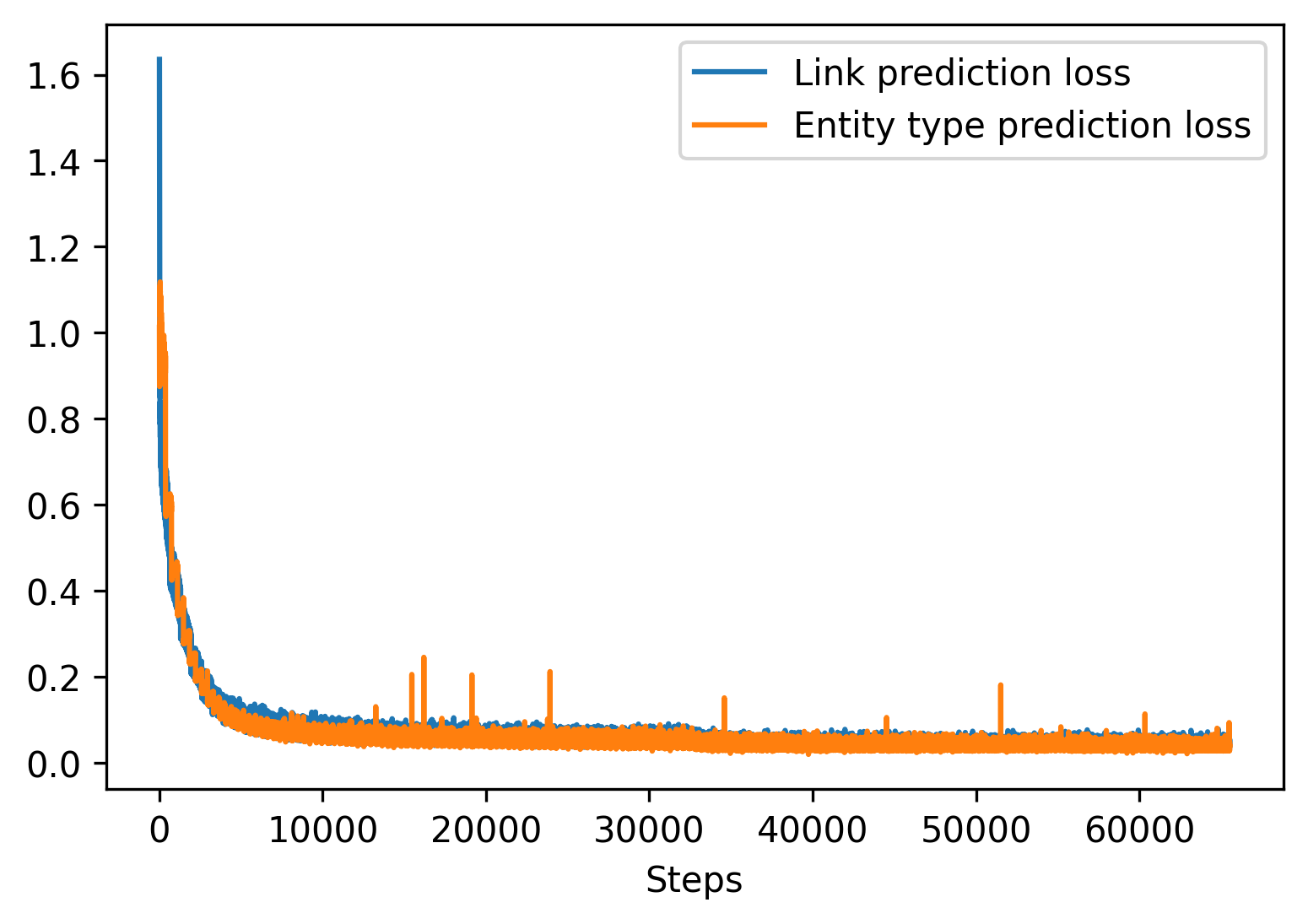} &
    \includegraphics[width=.32\linewidth]{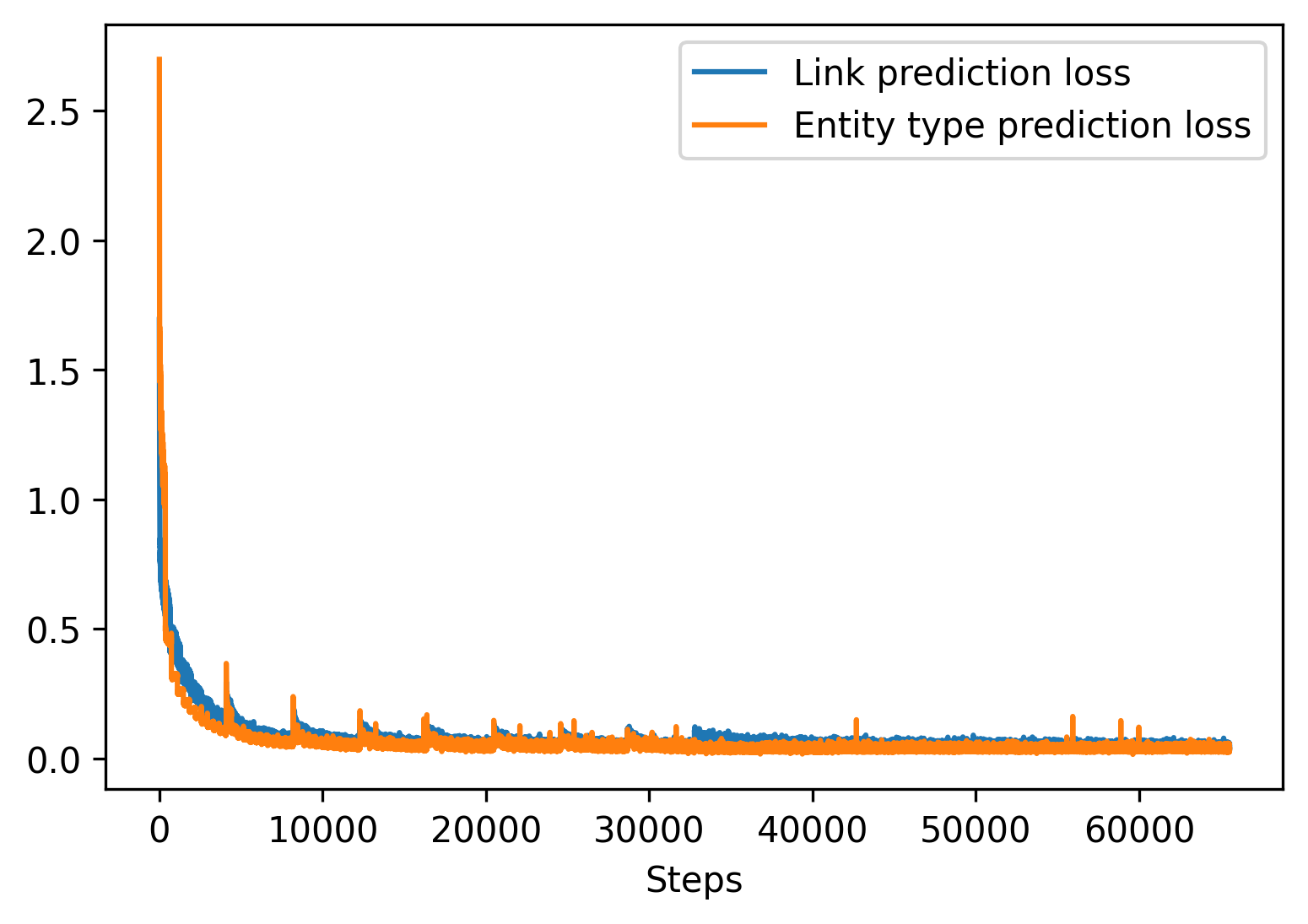} &
    \includegraphics[width=.32\linewidth]{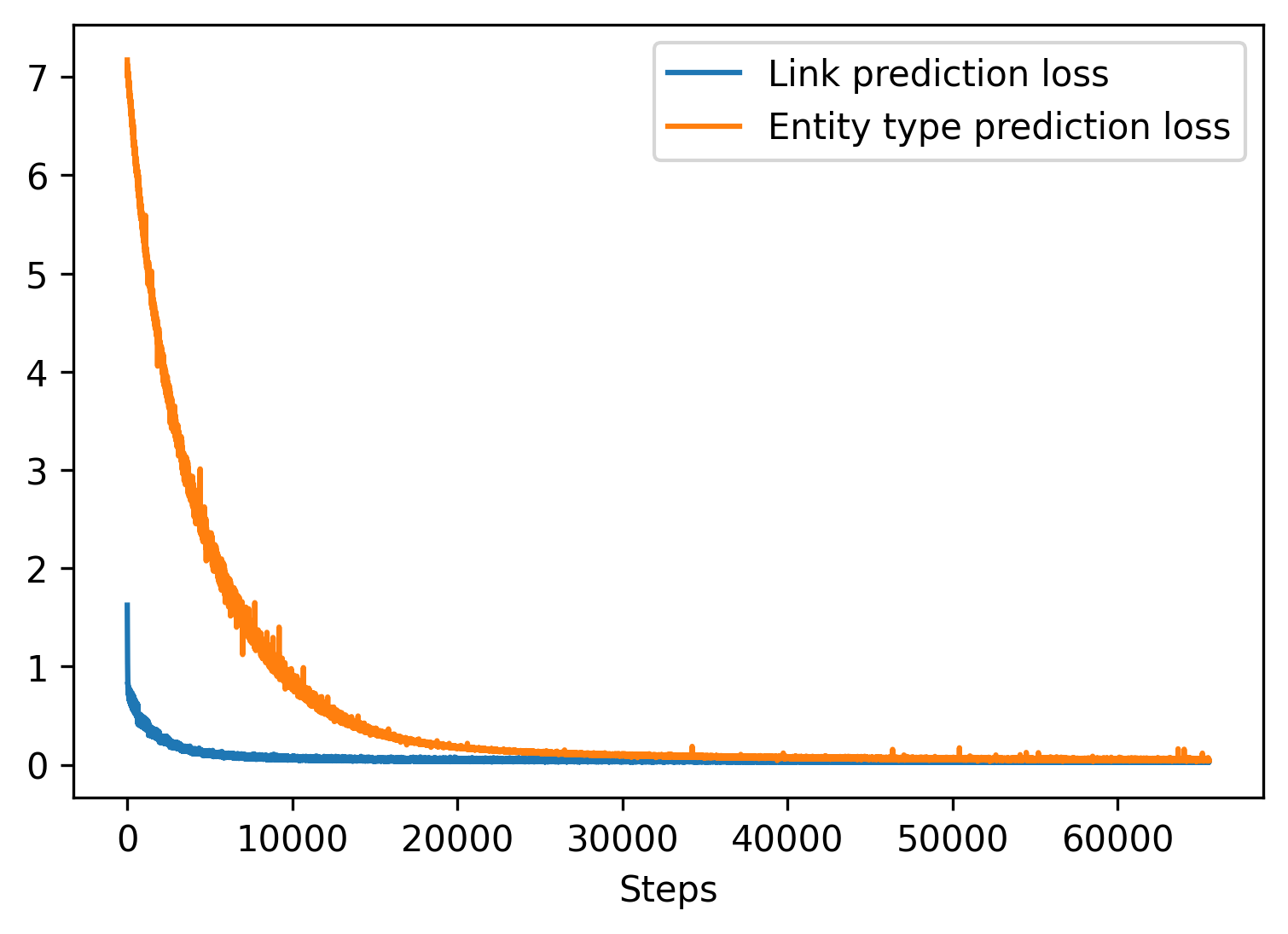}
    \\[\abovecaptionskip]
    \small (a) \# of alternating rounds = 256 & \small (b) \# of alternating rounds = 16 & \small (c) \# of alternating rounds = 1
  \end{tabular}
\caption{Training loss curves with respect to different 
numbers of alternating rounds.}\label{fig:loss_curves}
\end{figure*}

\begin{table*}
\centering
\caption{Inference time and memory complexity of KGET methods.}
\begin{tabular}{ l c c }
\toprule
Model & Inference time complexity & Memory complexity \\
\midrule
ConnectE \citep{zhao2020connecting}       & $O(T d_e^2 d_t)$ & $O((E + R)d_e + Td_t + d_e d_t)$\\
R-GCN \citep{schlichtkrull2018modeling}   & $O(TBL(E + T)d^2)$ & $O((E + R + T)d + BLd^2 + BLR)$\\
WGCN \citep{zhao2021wgcn}                 & $O(TL(E + T)d^2)$ & $O((E + R + T)d + Ld^2 + LR)$\\
CET \citep{pan2021context}                & $O(T(Td^2 + 2T))$ & $O((E + R + 3T)d)$\\
KGE Models                               & $O(Td)$ & $O((E + R + T)d)$ \\
\midrule
AsyncET (Ours) & $O(2Td)$ & $O((E + R + T + P)d)$ \\
\bottomrule
\end{tabular}
\label{tab:complexity}
\end{table*}

\subsection{Number of Alternating Rounds}\label{sec:alter_round}

In asynchronous learning, we alternate between the link prediction stage
(Stage 1) and the entity type prediction stage (Stage 2) after a few
stochastic gradient descent steps. The link prediction loss in Eq.
(\ref{equ:lp_loss}) and the entity type prediction loss in Eq.
(\ref{equ:tp_loss}) are minimized in Stage 1 and Stage 2, respectively.
The training process begins with Stage 1 and switches to Stage 2 after
$N_{s,1}$ stochastic gradient descent steps. Similarly, it conducts
$N_{s,2}$ stochastic gradient descent steps and then switches back to
Stage 1. We call one entire cycle of performing Stage 1 and Stage 2 once
an alternating round. In our experiments, we set $N_{s,1}=N_{s,2}=N_s$
and use $N_r$ to denote the number of alternating rounds.

We plot the MRR performance on YAGO43kET using TransE as the scoring
function as a function of $N_s$ and $N_r$ in Fig. \ref{fig:alternate} in
the green line.  We conduct experiments using $N_s$ steps, with $N_s=1,
16, 256, 4096, 65536$ in one round. Besides, we set $N_s \times N_r=65,536$.
The relation of $N_s$ and $N_r$ is shown by the red line. The smaller
$N_s$ being set means the stage alternation is more frequent. We see that the
performance is better when alternating between the two stages more
frequently. Clearly, asynchronous training contributes better entity and
type embedding quality when more frequent interactions exist
between the entity and type embedding spaces. 

We also plot the training loss curves as a function of alternating
rounds in Fig. \ref{fig:loss_curves}. It shows that the training loss
can be lower with more alternating rounds.  In addition, both the link
and entity type prediction loss are successfully
reduced during the training.  In other words, the two training stages
are mutually beneficial.  When alternating between the two stages only
once, as shown in Fig.  \ref{fig:loss_curves} (c), the loss curves go
down slower than the other two cases. Thus, more alternating rounds
help convergence. We conclude that alternating between two stages can
fine-tune entity and type embeddings to approach the global optimal in
optimization.

\subsection{Complexity Analysis} 

We conduct complexity analysis on the inference time and the number of
model parameters for several representative methods in Table
\ref{tab:complexity}, where $d$, $E$, $R$, $T$, and $P$ denote the
embedding dimension, numbers of entities, relations, entity types, and
auxiliary relations, respectively. For ConnectE, the embedding
dimensions for entities and types are denoted as $d_e$ and $d_t$,
respectively.  For GCN-based methods, $B$ is the number of bases to
decompose the propagation matrix in GCN layers, and $L$ is the number of
GCN layers.  We see that KGE methods are the most efficient methods in
terms of inference time complexity under the same embedding dimension.
ConnectE is also an embedding-based method, but it tries to learn a
matrix to connect the entity and type space.  Thus, its time complexity
and number of model parameters are proportional to $d^2$, which are
larger than KGE methods.  The complexity of GCN methods is correlated
with the number of layers and the number of nodes in the graph.  As a
result, the inference time complexity is proportional to $(E + T)Ld^2$,
which is highly inefficient in testing. The complexity of AsyncET is
similar to KGE methods, except it needs additional model parameters to
store embeddings for auxiliary relations.  For time complexity, AsyncET
requires twice the inference time as that of KGE methods since it considers
joint probabilities $f(e, p', t')$ in candidate ranking.  KGE methods
only need to calculate the conditional probabilities $f(e, t' \mid
\textit{hasType})$. 

\begin{table}
\centering
\caption{Top 3 predicted entity types by AsyncET for entities in YAGO43kET. Groundtruth is marked in boldface.}
\begin{tabular}{ c c }
\toprule
Entity & Top 3 Type Predictions \\
\midrule
\multirow{3}{*}{Mihail Majearu} & \textbf{Romanian footballers} \\
  & Alkmaar players \\
  & People from Glasgow \\
\midrule
\multirow{3}{*}{David Cross} & \textbf{Jewish actors} \\
   & 21st-century American actors \\
   & American humorists \\
\midrule
\multirow{3}{*}{Zhejiang} & Cities in Zhejiang \\
   & Provincial capitals in China \\
   & \textbf{Administrative divisions of China} \\
\bottomrule
\end{tabular}
\label{tab:example}
\end{table}

\subsection{Qualitative Analysis}

We show some examples of predicted types in Table \ref{tab:example}. 
In all three examples, the groundtruth ranks among the top three. 
In the first example, for entity \emph{Mihail Majearu}, the model can 
successfully rank the groundtruth at the top one. In addition, the top three
predicted types are all persons, and the 2nd prediction is also related
to football. In the second example, for entity \emph{David Cross}, who is
a comedian, the model can rank the groundtruth at the top one again.
The other two entity types are also relevant to the entity. They are 
valid types. In the last example, for entity \emph{Zhejiang}, although
the groundtruth only ranks as the third, the first two predicted types
are both relevant to the entities.  Note that \emph{Zhejiang} is a
province instead of a city in China.  The granularity of the entity is
not predicted correctly in the top two choices. 

\section{Conclusion and Future Work} \label{sec:conclusion}

Multiple auxiliary relations were proposed to solve the KGET task in
this work. Three methods for the design of auxiliary relations were
compared. The efficient assignment is recommended among the three since
it is scalable to datasets containing many entity types. In addition,
asynchronous embedding learning was proposed to achieve better
entity and type embeddings by alternatively predicting missing links and types
.  It was shown by experimental results that AsyncET
outperforms SOTA in H@3 and H@10 with much lower inference complexity
and fewer model parameters.  As future extensions, we will investigate
how the sparsity of the typing information affects the performance of
KGET methods.  
We aim to not only develop a time- and parameter-efficient model,
but achieve less performance degradation when 
trained with fewer labeled entity types. 

\bibliographystyle{plainnat}
\renewcommand\refname{Reference}
\bibliography{main}

\end{document}